\documentclass[10pt,twocolumn,letterpaper]{article}

\usepackage{iccv}
\usepackage{times}
\usepackage{epsfig}
\usepackage{graphicx}
\usepackage{amsmath}
\usepackage{amssymb}
\usepackage[dvipsnames]{xcolor}

\usepackage{amsmath,graphicx}
\usepackage{booktabs}
\usepackage{algorithm}
\usepackage{algpseudocode}
\usepackage{lipsum}

\usepackage{authblk}

\usepackage[breaklinks=true,bookmarks=false]{hyperref}

\iccvfinalcopy 

\def\iccvPaperID{RCV-76} 

\ificcvfinal\fi

\begin{document}

\title{Extending TrOCR for Text Localization-Free OCR \\
of Full-Page Scanned Receipt Images\vspace*{-0.2cm}}


\author{Hongkuan Zhang$^{1}$, Edward Whittaker$^{2}$, Ikuo Kitagishi$^{3}$ \\ $^{1}$ Nagoya University \ $^{2}$ K.K. Best Path Research \ $^{3}$ Money Forward, Inc.\vspace*{-0.3cm}}

\affil{{\tt\small zhang.hongkuan.k5@s.mail.nagoya-u.ac.jp$^{1}$}\\{\tt\small ed@bestpathresearch.com$^{2}$} \ {\tt\small kitagishi.ikuo@moneyforward.co.jp$^{3}$}\vspace*{-0.6cm}}



\maketitle
\ificcvfinal\thispagestyle{empty}\fi

\begin{abstract}
   Digitization of scanned receipts aims to extract text from receipt images and save it into structured documents. This is usually split into two sub-tasks: text localization and optical character recognition (OCR). Most existing OCR models only focus on the cropped text instance images, which require the bounding box information provided by a text region detection model. Introducing an additional detector to identify the text instance images in advance adds complexity, however instance-level OCR models have very low accuracy when processing the whole image for the document-level OCR, such as receipt images containing multiple text lines arranged in various layouts. To this end, we propose a localization-free document-level OCR model for transcribing all the characters in a receipt image into an ordered sequence end-to-end. Specifically, we finetune the pretrained instance-level model TrOCR with randomly cropped image chunks, and gradually increase the image chunk size to generalize the recognition ability from instance images to full-page images. In our experiments on the SROIE receipt OCR dataset, the model finetuned with our strategy achieved 64.4 F1-score and a 22.8\% character error rate (CER), respectively, which outperforms the baseline results with 48.5 F1-score and 50.6\% CER. The best model, which splits the full image into 15 equally sized chunks, gives 87.8 F1-score and 4.98\% CER with minimal additional pre or post-processing of the output. Moreover, the characters in the generated document-level sequences are arranged in the reading order, which is practical for real-world applications.
\end{abstract}

\section{Introduction}

Scanned receipt digitization aims at documenting text in receipts. This process was formally defined as Scanned Receipts OCR and Information Extraction (SROIE) task in the ICDAR 2019 competition \cite{huang2019icdar2019}, which provides a benchmark dataset, called SROIE, and splits the task into localization and recognition sub-tasks. Existing works focus on either the text detection \cite{wang2019efficient,wang2020deep,liao2020real} or character recognition \cite{shi2016end,baek2019wrong,yu2020towards}, and most OCR models can only transcribe text from cropped text instance-level images as opposed to document-level receipt images. Introducing an additional detector to identify the text instance images in advance increases system complexity, and it requires post-processing to combine the instance-level sequences to obtain a document-level transcription. To this end, finetuning an instance-level OCR model to generalize its recognition ability to full page images could be more efficient for document-level OCR, while maintaining the same accuracy.

\begin{figure}[t]
\centering
\includegraphics[width=\columnwidth]{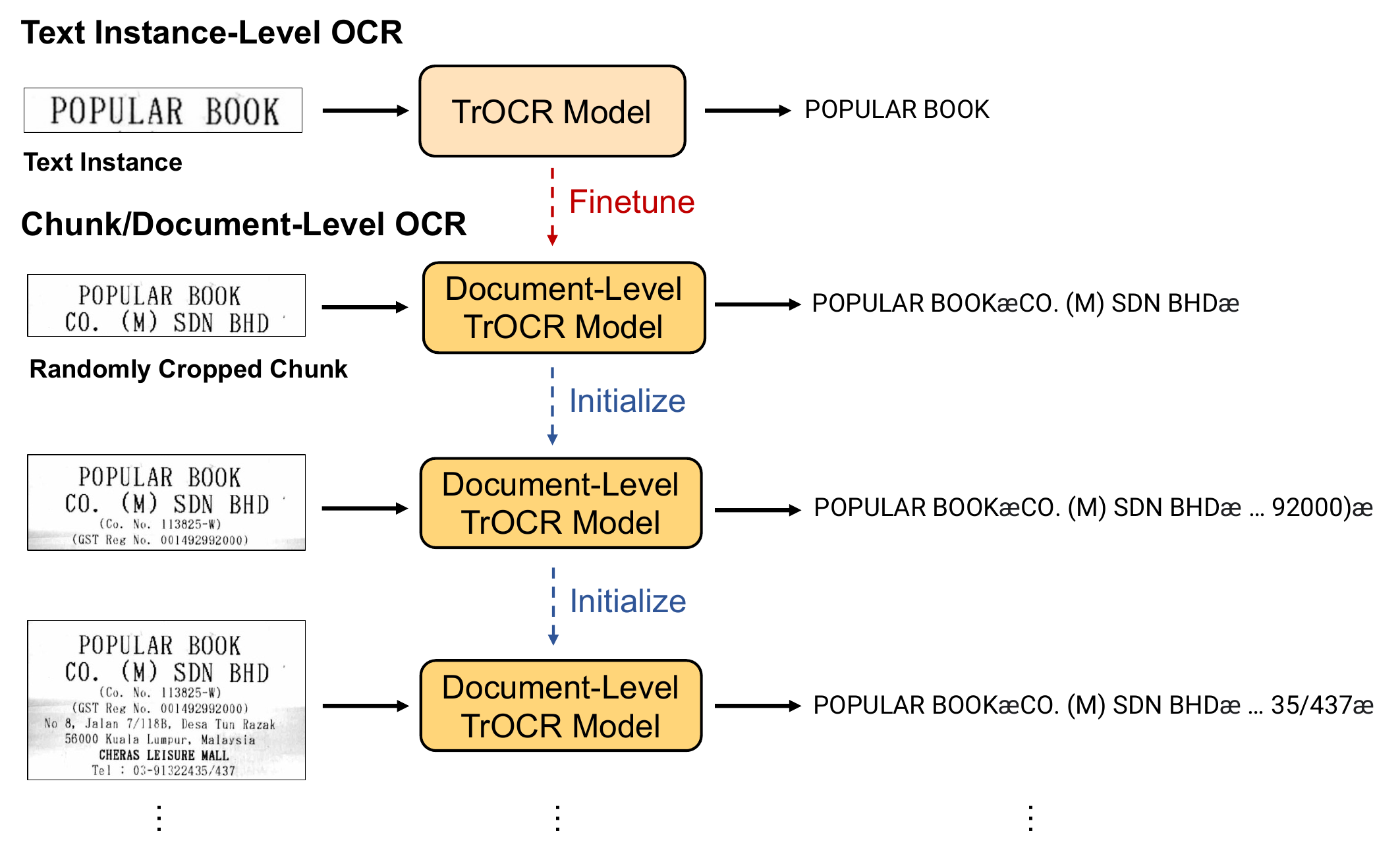}\caption{Our proposed step-by-step finetuning strategy for adapting TrOCR to the document-level OCR. ``æ'' indicates the separator token for dividing characters in each text-line.}
\label{introduction-figure}
\vspace{-1em}
\end{figure}

Recently, a pretrained Transformer-based OCR model TrOCR \cite{li2023trocr} was proposed which achieved state-of-the-art performance on the SROIE dataset. However, TrOCR was trained using only text instance images, which makes its adaptation to document-level OCR challenging because of the great variation in input images, which include many more characters and more lines than the single lines it was trained on. To explore the potential of TrOCR for end-to-end document-level OCR without text localization, we propose an efficient step-by-step finetuning strategy as shown in Fig.\ref{introduction-figure}. Specifically, we first randomly split the whole receipt image into image chunks whose size is closer to the original text instance images. These are then used to finetune the TrOCR model for what we call ``chunk-level'' OCR, and we gradually increase the chunk size to introduce more difficult chunks containing more lines and characters. Every time we use the model finetuned in the previous step to initialize the current model. Finally, we train using the entire, un-chunked, receipt images to achieve document-level OCR. The intuition behind this strategy is to progressively get the model to generalize its recognition ability to larger images. We define the order of characters in the chunk-level label as top-left to bottom-right, and propose a method to construct the reference label for each chunk automatically. We also include a text-line separator token in the constructed reference label, which aims to encode the line segmentation for the layout learning and also facilitate post-processing the model output into lines during inference.

We conduct experiments on the SROIE dataset for the document-level OCR. We cannot directly compare our results with those in the literature, since they only focus on instance-level OCR. Thus we construct two baseline finetuning methods for comparison. The experimental results show that our method achieves better performance than the two baselines on both word-level and character-level metrics. We finetuned TrOCR as well as the document understanding model Donut \cite{kim2022donut} for comparison. Both models have different input sizes, and using our method we expect we could eventually find an optimal input size in terms of accuracy and computational efficiency. The main contributions of our work are summarized as follows:


\begin{figure*}[ht]
\centering
\includegraphics[width=\textwidth]{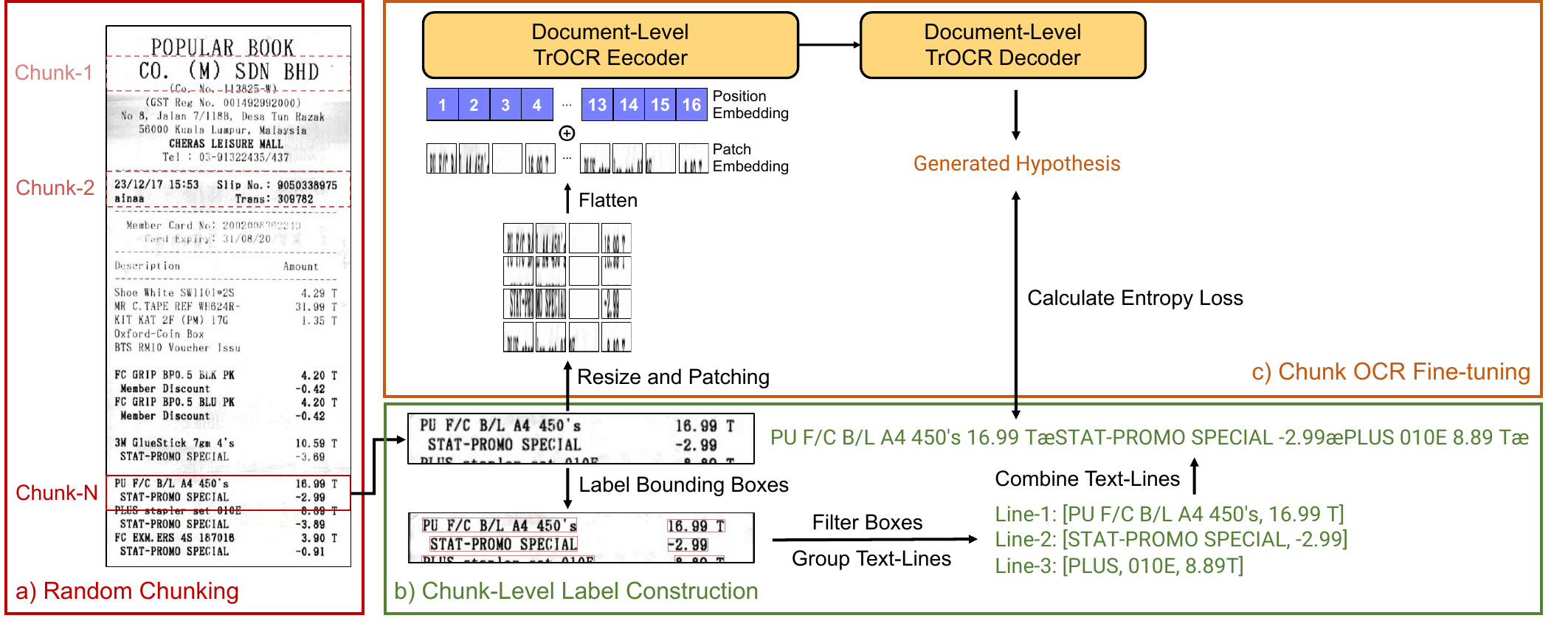}
\caption{Finetuning the TrOCR model for chunk-level OCR.}
\label{chunk-ocr}
\vspace{-0.4em}
\end{figure*}

\begin{itemize}
    \setlength\itemsep{0.1em}
    \item We propose a method to construct chunk-level reference labels automatically using only annotated instance-level labels. This method can be easily applied to other OCR datasets.
    \item The generated characters are arranged in the reading order with a unique text-line separator token for post-processing, which is practical for real-world applications.
    \item We propose a step-by-step finetuning strategy to adapt TrOCR for document-level OCR, which can process the entire image and achieve competitive performance.
\end{itemize}

\section{Related Works}
\label{sec:related-works}

\noindent\textbf{CNN-Based OCR models} Based on our target, we mainly focus on end-to-end OCR models which include two modules for detection and recognition, respectively. Previous research treats these two problems independently \cite{alsharif2013end,wang2012end,wang2011end} by combining a text detector with a recognition model. Since the interaction between the two modules can complement each other to avoid the error propagation, recent research \cite{busta2017deep,he2018end,liu2020abcnet,jaderberg2014deep} jointly optimizes the two modules by sharing the intermediate results. However, all these models include a text detection module explicitly or implicitly whereas we aim to perform document-level OCR without extracting intermediate text regions.



\noindent\textbf{Transformer-Based OCR models} A Transformer-based model TrOCR was proposed recently for the receipt OCR task. TrOCR incorporates a vision transformer and a language model in its encoder-decoder architecture, which was trained on large-scale printed and handwritten OCR data for robust text recognition. Several Transformer-based models were also proposed focusing on hand written text OCR \cite{wick2021transformer,strobel2022transformer} or scene text OCR \cite{atienza2021vision,tang2021visual,orihashi2021utilizing}. However, all these models are restricted to the transcription of the cropped text-line or instance images instead of the full images. Recently, an OCR-free document understanding model Donut \cite{kim2022donut} has been proposed lately, which includes a pseudo-OCR pretraining task to transcribe texts from document images.

\section{Methodology}
\label{sec:methodology}
\subsection{TrOCR Model Architecture}
We will first introduce TrOCR as the backbone model for our finetuning work. TrOCR is a Transformer-based OCR model which consists of a pretrained vision Transformer encoder BEiT \cite{bao2021beit} and a pretrained language model decoder RoBERTa \cite{liu2019roberta} as shown in Fig.\ref{chunk-ocr}. To recognize characters in the cropped text instance images, the images are first resized into square boxes of size $384\times384$ pixels and then flattened into a sequence of 576 patches, which are then encoded by BEiT into high-level representations and decoded by RoBERTa into corresponding characters step-by-step. TrOCR was pretrained with 684M textlines in English, which ensures the robust recognition ability for characters in various formats. However, TrOCR can only handle cropped single-line text instance images, which leads to underperformance when finetuning the model directly using whole receipt images. Therefore, a better finetuning strategy to adapt the model recognition from cropped images to full images is required.

\subsection{Chunk-Level OCR Finetuning}
To leverage TrOCR for whole image text recognition, we propose to finetune the model with image chunks extracted from the whole images for chunk-level OCR. As Fig.\ref{chunk-ocr} shows, our finetuning pipeline contains three modules: (i) randomly sample image chunks from the full receipt image; (ii) construct the label for the sampled chunks; and (iii) finetune the model with chunks and corresponding chunk labels. We will introduce the random sampling and the finetuning process in this section. 


Randomly sampling image chunks from whole images aims to obtain larger images for training the model. The reason for introducing randomness is that we hope to extract different chunks from each image across different epochs, which can increase data diversity and improve model generalization. Formally, to sample a chunk from a receipt image of width $W$ and height $H$, we first set a hyper-parameter $L$ for the chunk numbers that we will split a receipt image into, then define the image chunk size whose width $w$ is always the same as the corresponding image width $W$ and height $h$ equal to $H/L$. With the determined chunk size, we randomly select an image chunk starting point $s$ on the y-axis whose value ranges from $0$ to $H-H/L$, and crop the chunk between $s$ and $s+h$ on the y-axis. Last, we repeat the sampling $N$ times to extract multiple chunks from each receipt image. 


With the randomly cropped chunk images, we can obtain corresponding chunk-level labels with the method that will be described in the next section, and use the chunks and labels for chunk-level OCR finetuning. The chunk size determines the contents in the chunk which in turn affect the learning difficulty. Therefore we feed the model with smaller chunks whose resolution is similar to the text instance images at the early stage, and increase the chunk size gradually to finetune the model with progressively more difficult image chunks. Concretely, we start by setting $L$ with a large value which produces smaller chunks, then after the training is finished using the current $L$, we start the next stage training with smaller $L$, and every time we use the best model checkpoint from the previous training step to initialize the model in the current step. Finally, we increase the chunk size to the full image size (L=1) for document-level OCR. We use the notation Growing-Finetune for our proposed method.

\begin{algorithm}[t]
  \begin{algorithmic}[1]
    \Statex \textbf{Input:} overlapping threshold $\theta$, merging threshold $\delta$, chunk numbers to split \textit{L}, chunk numbers to sample \textit{N}
    \Statex \textbf{Output:} chunk set $X$ and chunk label set $Y$
    \Statex \textbf{Data:} input image $\textit{V} = \textit{W} \times \textit{H}$
    \State \textbf{Init} $X\leftarrow \{\}$, $Y\leftarrow \{\}$, $n\leftarrow1$, $h \leftarrow H/L$
    \For{$n$ = $1$ to $N$}
    \State Randomly select a starting point $s \in (0,H-H/L)$
    \State Chunk $x$ between $s$ and $s+h$
    \State Gather boxes $B=\{b_{i}\}_{i=1}^{I}$ and labels $R=\{r_{i}\}_{i=1}^{I}$
    \State Sort boxes and labels by y-axis values of anchors
    \For{$b_{i}$ in B} \Comment{filtering boxes}
        \If{$overlap(b_{i},x) \leq \theta$}
        \State Remove $b_{i}$ from $B$
        \EndIf
    \EndFor
    \State \textbf{Init} merged boxes $B^{'} \leftarrow \{\}$ and labels $T^{'} \leftarrow \{\}$
    \For{$b_{i}$ in $B$} \Comment{merging text-line labels}
        \If{$b_{i} \notin B{'}$}
            \State \textbf{Init} text-line label set $t_{i} \leftarrow \{r_{i}\}$
            \For{$b_{j}$ in $B-\{b_{i}\}$}
                \If {$v\_overlap(b_{i},b_{j}) \geq \delta$}
                    \State Add $r_{j}$ to $t_{i}$
                \EndIf
            \EndFor
            \State Sort $t_{i}$ based on x-axis values of anchors
            \State $t_{i} \leftarrow$ Concat labels in $t_{i}$ with whitespace
            \State Add $t_{i}$ to $T^{'}$
            \State Add $b_{i}$ to $B^{'}$
        \EndIf
    \EndFor
    \State $y \leftarrow$ Concat labels in $T^{'}$ with the separator æ
    \State Add $x$ to $X$
    \State Add $y$ to $Y$
    \EndFor
\State \textbf{return} $X,Y$
  \end{algorithmic}
  \caption{Chunk-Level Label Construction}\label{chunk-label-algo}
\end{algorithm}

\subsection{Chunk-Level Label Construction} 



Chunks usually contain multiple text-lines and some characters are split vertically in half as shown in Fig.\ref{chunk-ocr}. Therefore, to construct labels for randomly cropped chunks, a definition for the character order in the label and which split characters should be included in the label is required. In this paper, we define the top-left to bottom-right reading order for characters in the chunk as the correct order, and set a overlapping threshold $\theta$ to include characters with an overlapping rate larger than $\theta$ to make sure no unrecognizable split characters are mistakenly included in the label. Based on these definitions, we construct the chunk-level label with the use of annotated texts and corresponding bounding boxes information as shown in Algorithm \ref{chunk-label-algo}. We first gather the $I$ bounding boxes in the randomly cropped image chunk and filter out boxes with overlapping rate less than $\theta$, and sort boxes as well as labels based on the y-axis values of the left-upper anchors of boxes. To align boxes horizontally in the same line, we define a merging threshold $\delta$ to merge boxes that overlap vertically over the threshold into text-line level labels, and sort the boxes in each group based on the x-axis values of the left-upper anchors from left to right. Lastly, we concatenate labels of boxes in each group with white space as text-line labels, which are concatenated with text-line label separator token ``æ". This character does not appear in any of the receipts and is used to encode the line segmentation for the receipt layout learning. To clarify, we only use the boxes for the label construction and no box is used during inference.

\section{Experiments}

\subsection{Settings} 
\noindent\textbf{Dataset} We use the SROIE dataset from the ICDAR 2019 competition for our experiments. The SROIE OCR task focuses on text recognition of cropped receipt images. There are 626 and 361 images in the training and testing set, respectively, which are annotated with ground truth bounding boxes and corresponding texts, and we keep the train and test data split the same as the TrOCR setting. We randomly sample 60 images from the training set to construct the validation set. For the chunk-level OCR setting, the training chunks are randomly sampled from each image in the training set, while the validation and testing chunks are sequentially cropped from each image to ensure the full image area is covered for the evaluation. All the labels are obtained by the chunk-level label construction method in Section 3.3.

\noindent\textbf{Hyper-parameters} We use \texttt{trocr-base-printed}\footnote{https://huggingface.co/microsoft/trocr-base-printed} model checkpoint finetuned with the original SROIE training data from Hugging Face \cite{wolf-etal-2020-transformers} for our own finetuning experiments. The boxes filtering threshold $\theta$, text-line merging threshold $\delta$ and sampled chunk number $N$ for training data are 0.3, 0.5, and 20, respectively, which are determined by results on the validation set. For the split chunk number $L$, we first compute the distribution of text-line numbers of training images as shown in Fig.\ref{distribution-figure}, then use the median number 30 as the initial value. By decreasing the value to 15, 7, 4, 2 and 1, we train the model with increasingly larger chunks at subsequent stages. We set the beam search size as 5 for the text generation. We also finetune the Donut model (\texttt{donut-base}\footnote{https://huggingface.co/naver-clova-ix/donut-base}) which has been pretrained to read texts from a large number of document images for comparison.


\noindent\textbf{Evaluation Metrics} We use two evaluation metrics adopted in OCR tasks to evaluate the performance: Word-Level precision, recall, F1 (Word-Level PRF) and Character Error Rate (CER). The Word-Level PRF focuses on correctly matched words in the hypothesis without considering the word order, while the CER focuses on the character-level substitutions, deletions, and insertions as:
\begin{figure}[t]
\centering
\includegraphics[width=0.82\columnwidth]{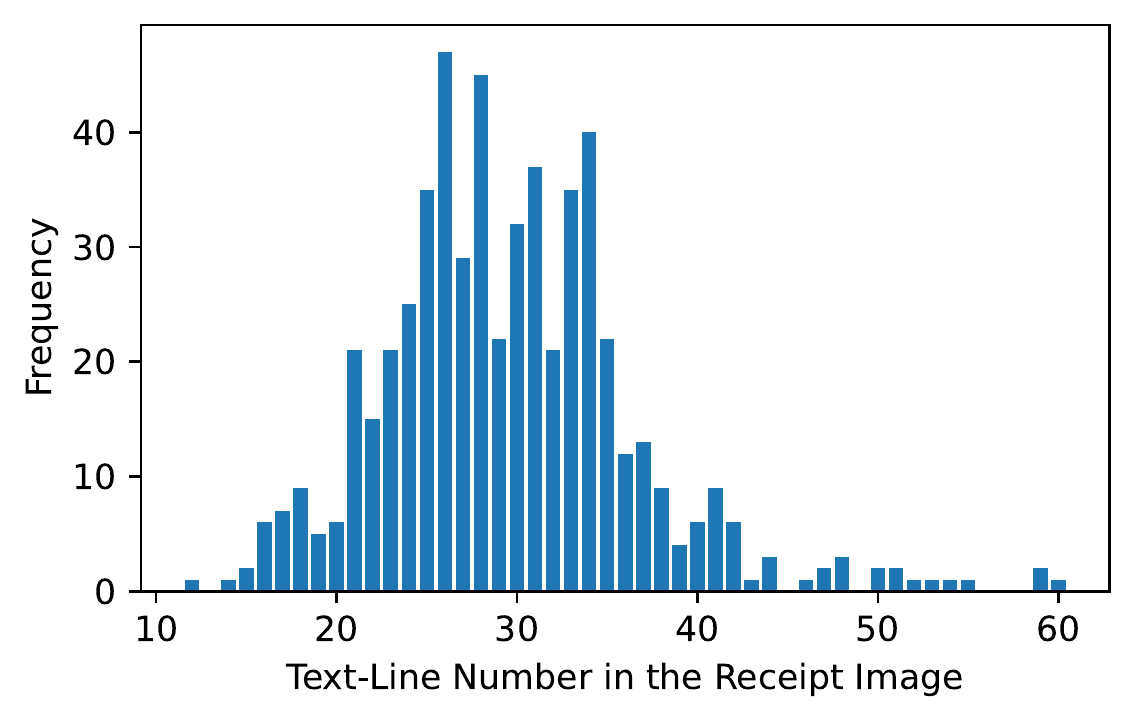}
\caption{Distribution of the number of text lines in receipt images.}
\label{distribution-figure}
\vspace{-0.4em}
\end{figure}
$$
CER = (S + D + I) / (S+D+C)
$$

\noindent where $S$, $D$, $I$, and $C$ are the number of substitutions, deletions, insertions, and correct characters, respectively.


\subsection{Baselines}

\noindent\textbf{Direct-Finetune} This is a straight-forward method that uses the whole receipt image for the document-level OCR directly. Concretely, we resize the whole image as input and split it into patches, and finetune the model with patches and document-level labels end-to-end. Resizing the whole image to $384\times384$ pixels will cause a large variation in the average resolution of each character for the recognition.

\noindent\textbf{Concatenate-Finetune} This is a compromise strategy to keep the input resolution closer to the original setting, which splits the image equally into several chunks, and embeds each chunk into sequences that are concatenated in the temporal dimension to construct document-level inputs. Since the linearly increased input length brings higher computational cost, we restrict the split number to 4 and interpolate the position embeddings to adjust for longer inputs.

\begin{figure*}[t]
\centering
\includegraphics[width=0.75\textwidth]{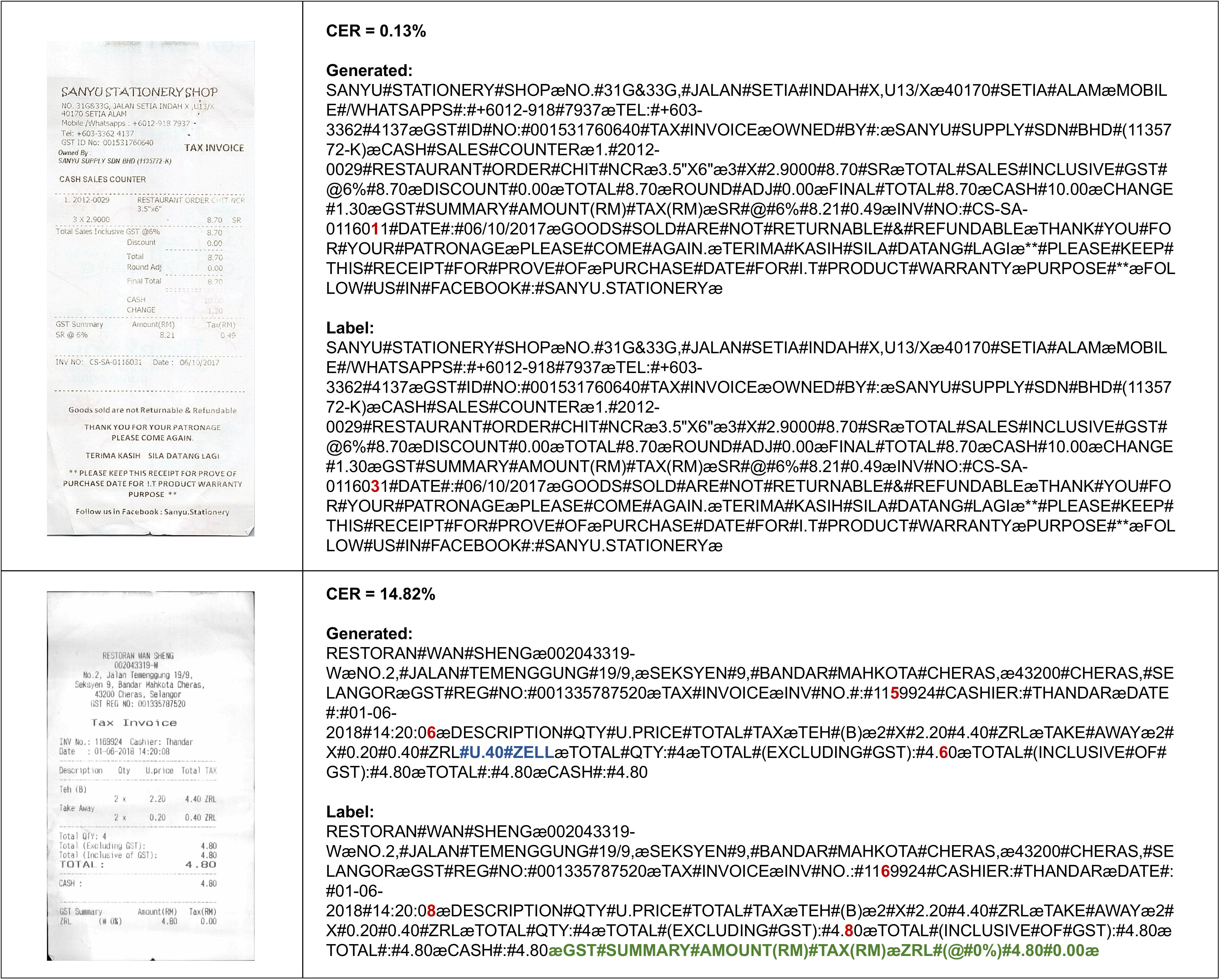}
\caption{Error analysis on the generated full-page receipt OCR texts. Characters in the red, blue, and green colors represent the substitution, insertion, and deletion errors, respectively. Black characters are the corresponding correct content.}
\vspace{-0.2cm}
\label{error-analysis}
\end{figure*}

\subsection{Quantitative Analysis}
\label{sec:experiments}
\noindent\textbf{Performance Comparison} We first compare the performance of our method and two baselines. As the results in Table \ref{general-performance} show, our method outperforms the two baselines on all metrics, which demonstrates the improved generalization ability for full-page images with Growing-Finetune. Moreover, the worst performance with Concatenate-Finetune also highlights the increased computational complexity on the longer inputs. Compared with the original TrOCR model, our results are worse since the evaluation for the longer and ordered document-level sequence is more strict, which also reveals the efficiency of our method to achieve good results without using the bounding box information for the localization. Furthermore, when applying our method to Donut for comparision, we see that the model achieves similar performance to the Direct-Finetune, which indicates that Growing-Finetune is more effective when the discrepancy of input image resolution between pretraining and finetuning is large. 

\begin{table}[t]
\centering
\resizebox{\columnwidth}{!}{%
\begin{tabular}{@{}lcccc@{}}
    \toprule
    Model & Precision & Recall & F1 & CER (↓) \\
    \midrule
    \textit{Instance-Level OCR} & & & & \\
    TrOCR & 96.1 & 96.2 & 96.2 & 0.95 \\\midrule
    \textit{Document-Level OCR w/ TrOCR} & & & & \\
    TrOCR+Direct-Finetune & 51.9 & 45.4 & 48.5 & 50.6 \\
    TrOCR+Concatenate-Finetune & 23.5 & 47.2 & 31.3 & 122.5 \\
    TrOCR+Growing-Finetune (Ours) & \textbf{66.4} & \textbf{62.5} & \textbf{64.4} & \textbf{22.8} \\
    \midrule
    \textit{Document-Level OCR w/ Donut} & & & & \\
    Donut+Direct-Finetune & \textbf{91.8} & \textbf{91.4} & \textbf{91.6} & \textbf{3.1} \\
    Donut+Growing-Finetune (Ours) & 91.3 & 91.0 & 91.1 & 3.5 \\
    \bottomrule
\end{tabular}}
\caption{Model performance comparison with different finetuning strategies for whole document OCR.}
\label{general-performance}
\end{table}

\begin{figure*}[t]
\centering
\includegraphics[width=0.75\textwidth]{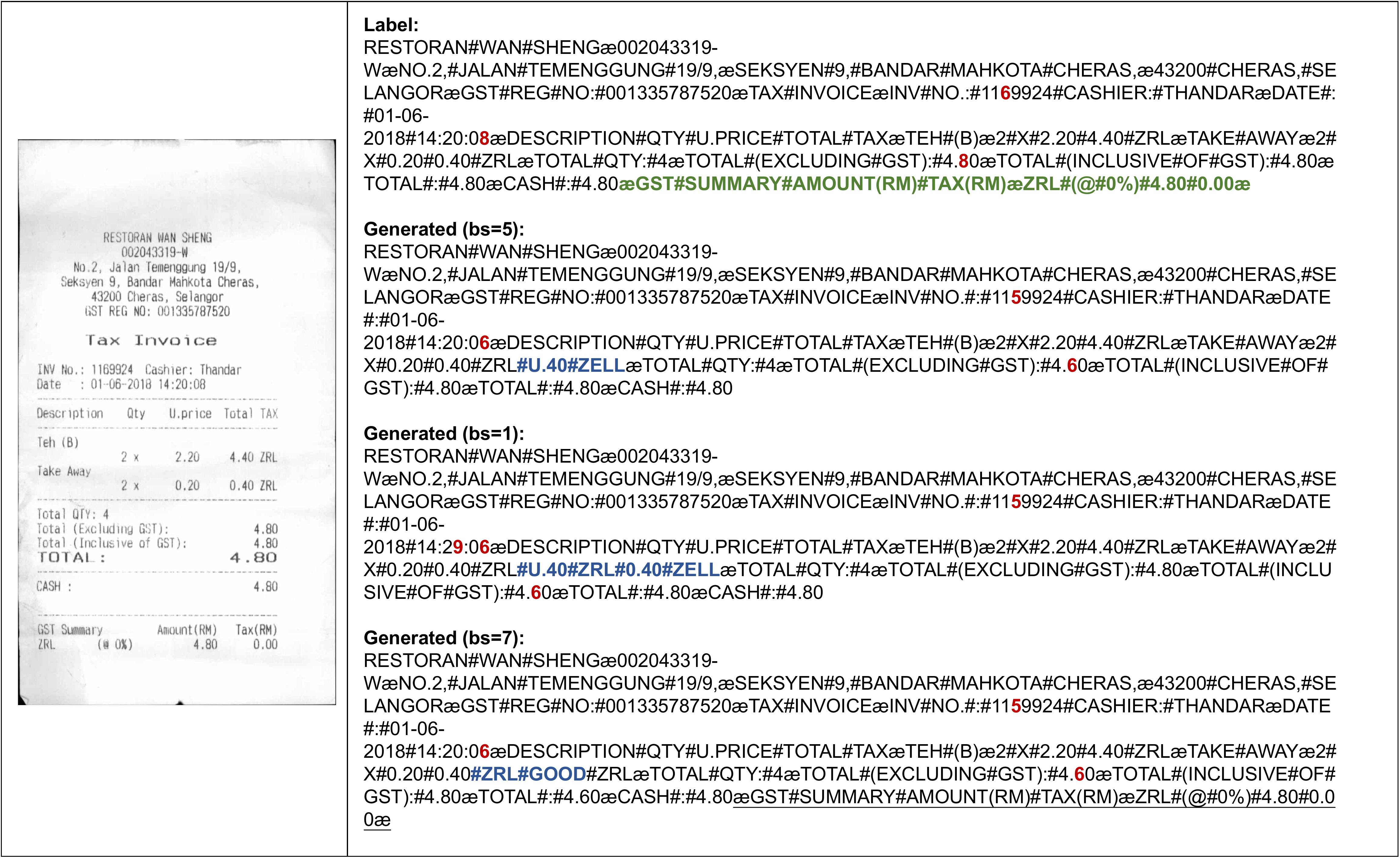}
\caption{Generated Texts with Different Beam Search Size (bs). The underlined texts are deleted contents with the default beam size.}
\label{beam-size-analysis}
\end{figure*}

\begin{table}[t]
\centering
\resizebox{\columnwidth}{!}{%
\begin{tabular}{@{}ccccccccc@{}}
    \toprule
    L & Precision & Recall & F1 & CER & S & I & D & Avg. length \\
    \midrule
    30 & 87.4 & 87.2 & 87.3 & 6.88 & 5315 & 7379 & 8761 & 5.9 \\
    15 & \textbf{87.9} & \textbf{87.7} & \textbf{87.8} & \textbf{4.98} & \textbf{3835} & 4593 & \textbf{5256} & 9.6  \\
    7 & 84.6 & 83.8 & 84.2 & 5.64  & 5468 & 3076 & 5822 & 18.2 \\
    4 & 82.6 & 81.9 & 82.2 & 6.04  & 6542 & \textbf{2874} & 5416 & 30.1 \\
    2 & 77.5 & 76.6 & 77.0 & 9.51 & 13057 & 3992 & 8747 & 57.5 \\
    1 & 66.4 & 62.5 & 64.4 & 22.8 & 27450 & 6489 & 20400 & \textbf{107.5} \\
    \bottomrule
\end{tabular}}
\caption{TrOCR model performance with different chunk numbers per image. Larger number L indicates smaller chunks on average.}
\vspace{-0.4cm}
\label{different-chunk-numbers}
\end{table}

\noindent\textbf{Chunk Size Analysis} We then analyzed how the TrOCR model performance changes according to the input chunk size. Since the chunk size is determined by the chunk number $L$, we test the performance under different $L$ and the variation is shown in Table \ref{different-chunk-numbers}. We observed that performance improved with increased chunk size at first, since the larger chunk size will introduce fewer split characters which reduces errors caused by the insertion and deletion, but the performance decreases when the size becomes larger, since images with more content and a longer target sequence significantly increases the learning difficulty, especially for the half-page and full-page settings which doubles the error rates compared with their previous settings. The optimal trade-off is achieved at $L=15$, where each chunk contains roughly 2 text-lines, on average. It is worth noting that if the text detection in the baseline TrOCR were to miss 4\% of ground truth regions, our best method results would be better than the original TrOCR performance.


\noindent\textbf{Ablation Study} Lastly, we analyze the influence of two main factors in our strategy with the ablation study as shown in Table \ref{ablation-study}. By removing the separator token, we noticed the CER performance drop is minor. While by sampling chunks sequentially without randomness, we found the performance dropped significantly on all metrics, which indicates the importance of randomness in bringing more diverse data and larger data size among all epochs for the convergence. This may also be thought of as a form of data augmentation.

\begin{table}[t]
\centering
\resizebox{\columnwidth}{!}{%
\begin{tabular}{lcccc}
    \toprule
    Model & Precision & Recall & F1 & CER \\
    \midrule
    Growing-Finetune (L=15) & 87.9 & 87.7 & 87.8 & 4.98 \\
    \midrule
    -- Separator Token & 89.8 & 88.8 & 89.3 & 5.46 \\
    -- Random Sampling & 77.8 & 76.5 & 77.2 & 11.3 \\
    \bottomrule
\end{tabular}}
\caption{Ablation study for the factor influence analysis.}
\vspace{-0.2cm}
\label{ablation-study}
\end{table}


\subsection{Qualitative Analysis}
To understand the quality of generated document-level texts, we conduct the error analysis on generated full-page OCR texts with $L=1$ as shown in Fig.\ref{error-analysis}. For the first example, our model generates a much longer sequence than instance-level TrOCR and achieved 0.13\% error rate where only one character was mistakenly substituted. This indicates the model can generate high-quality texts for full-image receipts. On the other hand, for the second example, our model achieved a 14.82\% error rate and made substitution errors on numbers and insertion or deletion errors on sequences. We hypothesize that the insertions are caused by memorization during model training, and the deletions are caused by the small beam size. We generate texts for the second example with the beam size 1 and 7, respectively, and the results in Fig.\ref{beam-size-analysis} show that decoding with a larger beam size can successfully generate the missing contents which supports our hypothesis.

\section{Conclusions and Future Work}
\label{sec:conclusions}

In this paper, we propose a step-by-step finetuning strategy and an automatic label construction method for adapting TrOCR to perform document-level receipt image OCR. The finetuned model can handle the full image input and transcribe all characters into long ordered sequences. Moreover, our method outperforms other straight-forward finetuning baselines, which indicates the efficiency of finetuning with image chunks of increasing size. We observed the trade-off between performance and chunk size, and learned the importance of random sampling from the ablation study. Besides, our method is more effective in improving the generalization ability when the discrepancy of input image resolution between pretraining and finetuning is large, as demonstrated with experiments on the Donut model. We expect this study can serve as a baseline for future studies that concentrate on efficient finetuning methods for the document-level receipt OCR model construction.


\vfill\pagebreak

{\small
\bibliographystyle{ieee_fullname}

\begin{thebibliography}{10}\itemsep=-1pt

\bibitem{alsharif2013end}
Ouais Alsharif and Joelle Pineau.
\newblock End-to-end text recognition with hybrid hmm maxout models.
\newblock {\em arXiv preprint arXiv:1310.1811}, 2013.

\bibitem{atienza2021vision}
Rowel Atienza.
\newblock {Vision Transformer} for fast and efficient scene text recognition.
\newblock In {\em International Conference on Document Analysis and Recognition (ICDAR 2021)}, pages 319--334. Springer, 2021.

\bibitem{baek2019wrong}
Jeonghun Baek, Geewook Kim, Junyeop Lee, Sungrae Park, Dongyoon Han, Sangdoo Yun, Seong~Joon Oh, and Hwalsuk Lee.
\newblock What is wrong with scene text recognition model comparisons? dataset and model analysis.
\newblock In {\em Proceedings Of The IEEE/CVF International Conference on Computer Vision (ICCV 2019)}, pages 4715--4723, 2019.

\bibitem{bao2021beit}
Hangbo Bao, Li Dong, Songhao Piao, and Furu Wei.
\newblock {BEiT}: {BERT} pre-training of image transformers.
\newblock {\em arXiv preprint arXiv:2106.08254}, 2021.

\bibitem{busta2017deep}
Michal Busta, Lukas Neumann, and Jiri Matas.
\newblock Deep textspotter: An end-to-end trainable scene text localization and recognition framework.
\newblock In {\em Proceedings of the IEEE International Conference on Computer Vision (ICCV 2017)}, pages 2204--2212, 2017.

\bibitem{he2018end}
Tong He, Zhi Tian, Weilin Huang, Chunhua Shen, Yu Qiao, and Changming Sun.
\newblock An end-to-end textspotter with explicit alignment and attention.
\newblock In {\em Proceedings of the IEEE Conference on Computer Vision and Pattern Recognition (CVPR 2018)}, pages 5020--5029, 2018.

\bibitem{huang2019icdar2019}
Zheng Huang, Kai Chen, Jianhua He, Xiang Bai, Dimosthenis Karatzas, Shijian Lu, and CV Jawahar.
\newblock {ICDAR} 2019 competition on scanned receipt ocr and information extraction.
\newblock In {\em 2019 International Conference on Document Analysis and Recognition (ICDAR 2019)}, pages 1516--1520. IEEE, 2019.

\bibitem{jaderberg2014deep}
Max Jaderberg, Andrea Vedaldi, and Andrew Zisserman.
\newblock Deep features for text spotting.
\newblock In {\em European Conference on Computer Vision (ECCV 2014)}, pages 512--528. Springer, 2014.

\bibitem{kim2022donut}
Geewook Kim, Teakgyu Hong, Moonbin Yim, JeongYeon Nam, Jinyoung Park, Jinyeong Yim, Wonseok Hwang, Sangdoo Yun, Dongyoon Han, and Seunghyun Park.
\newblock Ocr-free document understanding transformer.
\newblock In {\em European Conference on Computer Vision (ECCV 2022)}, pages 498--517. Springer, 2022.

\bibitem{li2023trocr}
Minghao Li, Tengchao Lv, Jingye Chen, Lei Cui, Yijuan Lu, Dinei Florencio, Cha Zhang, Zhoujun Li, and Furu Wei.
\newblock {TrOCR}: Transformer-based optical character recognition with pre-trained models.
\newblock In {\em Proceedings of the AAAI Conference on Artificial Intelligence (AAAI 2023)}, volume~37, pages 13094--13102, 2023.

\bibitem{liao2020real}
Minghui Liao, Zhaoyi Wan, Cong Yao, Kai Chen, and Xiang Bai.
\newblock Real-time scene text detection with differentiable binarization.
\newblock In {\em Proceedings of the AAAI Conference on Artificial Intelligence (AAAI 2020)}, volume~34, pages 11474--11481, 2020.

\bibitem{liu2020abcnet}
Yuliang Liu, Hao Chen, Chunhua Shen, Tong He, Lianwen Jin, and Liangwei Wang.
\newblock {ABCNet}: Real-time scene text spotting with adaptive bezier-curve network.
\newblock In {\em proceedings of the IEEE/CVF Conference on Computer Vision and Pattern Recognition (CVPR 2020)}, pages 9809--9818, 2020.

\bibitem{liu2019roberta}
Yinhan Liu, Myle Ott, Naman Goyal, Jingfei Du, Mandar Joshi, Danqi Chen, Omer Levy, Mike Lewis, Luke Zettlemoyer, and Veselin Stoyanov.
\newblock {RoBERTa}: A robustly optimized bert pretraining approach.
\newblock {\em arXiv preprint arXiv:1907.11692}, 2019.

\bibitem{orihashi2021utilizing}
Shota Orihashi, Yoshihiro Yamazaki, Naoki Makishima, Mana Ihori, Akihiko Takashima, Tomohiro Tanaka, and Ryo Masumura.
\newblock Utilizing resource-rich language datasets for end-to-end scene text recognition in resource-poor languages.
\newblock In {\em ACM Multimedia Asia (MM Asia 2021)}, pages 1--5. ACM, 2021.

\bibitem{shi2016end}
Baoguang Shi, Xiang Bai, and Cong Yao.
\newblock An end-to-end trainable neural network for image-based sequence recognition and its application to scene text recognition.
\newblock {\em IEEE Transactions on Pattern Analysis and Machine Intelligence (TPAMI 2016)}, 39:2298--2304, 2016.

\bibitem{strobel2022transformer}
Phillip~Benjamin Str{\"o}bel, Simon Clematide, Martin Volk, and Tobias Hodel.
\newblock Transformer-based htr for historical documents.
\newblock {\em arXiv preprint arXiv:2203.11008}, 2022.

\bibitem{tang2021visual}
Xin Tang, Yongquan Lai, Ying Liu, Yuanyuan Fu, and Rui Fang.
\newblock Visual-semantic transformer for scene text recognition.
\newblock {\em arXiv preprint arXiv:2112.00948}, 2021.

\bibitem{wang2020deep}
Jingdong Wang, Ke Sun, Tianheng Cheng, Borui Jiang, Chaorui Deng, Yang Zhao, Dong Liu, Yadong Mu, Mingkui Tan, Xinggang Wang, et~al.
\newblock Deep high-resolution representation learning for visual recognition.
\newblock {\em IEEE Transactions on Pattern Analysis and Machine Intelligence (PAMI 2020)}, 43:3349--3364, 2020.

\bibitem{wang2011end}
Kai Wang, Boris Babenko, and Serge Belongie.
\newblock End-to-end scene text recognition.
\newblock In {\em 2011 International Conference on Computer Vision (ICCV 2011)}, pages 1457--1464. IEEE, 2011.

\bibitem{wang2012end}
Tao Wang, David~J Wu, Adam Coates, and Andrew~Y Ng.
\newblock End-to-end text recognition with convolutional neural networks.
\newblock In {\em Proceedings of the 21st International Conference on Pattern Recognition (ICPR2012)}, pages 3304--3308. IEEE, 2012.

\bibitem{wang2019efficient}
Wenhai Wang, Enze Xie, Xiaoge Song, Yuhang Zang, Wenjia Wang, Tong Lu, Gang Yu, and Chunhua Shen.
\newblock Efficient and accurate arbitrary-shaped text detection with pixel aggregation network.
\newblock In {\em Proceedings of the IEEE/CVF International Conference on Computer Vision (ICCV 2019)}, pages 8440--8449, 2019.

\bibitem{wick2021transformer}
Christoph Wick, Jochen Z{\"o}llner, and Tobias Gr{\"u}ning.
\newblock Transformer for handwritten text recognition using bidirectional post-decoding.
\newblock In {\em International Conference on Document Analysis and Recognition (ICDAR 2021)}, pages 112--126. Springer, 2021.

\bibitem{wolf-etal-2020-transformers}
Thomas Wolf, Lysandre Debut, Victor Sanh, Julien Chaumond, Clement Delangue, Anthony Moi, Pierric Cistac, Tim Rault, R{\'e}mi Louf, Morgan Funtowicz, et~al.
\newblock Transformers: State-of-the-art natural language processing.
\newblock In {\em Proceedings of the 2020 Conference on Empirical Methods in Natural Language Processing: System Demonstrations (EMNLP 2020)}, pages 38--45, 2020.

\bibitem{yu2020towards}
Deli Yu, Xuan Li, Chengquan Zhang, Tao Liu, Junyu Han, Jingtuo Liu, and Errui Ding.
\newblock Towards accurate scene text recognition with semantic reasoning networks.
\newblock In {\em Proceedings of the IEEE/CVF Conference on Computer Vision and Pattern Recognition (CVPR 2020)}, pages 12113--12122, 2020.

\end{thebibliography}

}

\end{document}